\documentclass[a4paper]{spie}  %>>> use this instead for A4 paper
\usepackage{url}
\usepackage[]{graphicx}

\usepackage{multirow}
\usepackage{rotating}
\newcommand{\mywidth}{0.9 \textwidth}
\graphicspath{{../figs/}{.}{./figs/}}
\title{Some observations on computer lip-reading: moving from the dream to the reality} 

\author{Helen L. Bear \supit{a}, Gari Owen\supit{b}, Richard Harvey\supit{a}, and Barry-John Theobald\supit{a}
\skiplinehalf
\supit{a}University of East Anglia, Norwich, NR4 7TJ, UK.\\
\supit{b}Annwvyn Solutions, Bromley, Kent, BR1 3DW, UK.
}
\vspace{-0.5cm}
\authorinfo{Further author information: (Send correspondence to RWH)\\RWH: E-mail: r.w.harvey@uea.ac.uk}
\begin{document} 
  \maketitle 

%%%%%%%%%%%%%%%%%%%%%%%%%%%%%%%%%%%%%%%%%%%%%%%%%%%%%%%%%%%%% 
\begin{abstract}
In the quest for greater computer lip-reading performance there are a number of tacit assumptions which are either present in the datasets (high resolution for example) or in the methods (recognition of spoken visual units called ``visemes'' for example).  Here we review these and other assumptions and show the surprising result that computer lip-reading is not heavily constrained by video resolution, pose, lighting and other practical factors.  However, the working assumption that visemes, which are the visual equivalent of phonemes, are the best unit for recognition does need further examination.  We conclude that visemes, which were defined over a century ago, are unlikely to be optimal for a modern computer lip-reading system.
\end{abstract}

%>>>> Include a list of keywords after the abstract 


%%%%%%%%%%%%%%%%%%%%%%%%%%%%%%%%%%%%%%%%%%%%%%%%%%%%%%%%%%%%%
\section{Background}
There has been consistent and sustained interest in building computer systems that can understand what humans are saying without hearing the audio channel \cite{bowden2013recent, cappelletta2012phoneme, davis1980,Bowden:2012fk}.  There are obvious applications for such systems in security but also in noisy environments such as cockpits, battlefields and crowds where audio recognition is likely to be impossible or highly degraded.  Early work consisted of very small vocabularies (often fewer than 10 words) \cite{Petajan1984:Automatic}, single speakers, high-definition video (often the camera would be zoomed into the lip region or the frame rate would be greater than 60 fields per second) \cite{Brooke1983:Analysis} and, often, the talker would wear special lipstick to allow easy segmentation and analysis of the lips \cite{kaucic1998accurate}.   Subsequently, our understanding of the problem has improved such that lip-reading in outdoor conditions (which requires very robust lip-tracking) and with 1000-voclabularies (which requires good machine learning) looks feasible.  The problem of speaker dependence is still only partially solved \cite{mvg:1763}.   One surprising recent result was a characterisation of the effect of resolution on lip-reading.  An informal understanding was that relatively high resolution was required (at least a couple of hundred pixels to span the lips).  In practice, it was reported in \cite{bearicip} that, provided the tracking was perfect, then fewer than 10 pixels can give acceptable results.  A further observation \cite{lan2012view} was that off-axis lip-reading gave slightly better  performance than full frontal (which is the default for most experiments).  It seems, when it comes to lip-reading, one's intuition might often be wrong -- indeed experimenters in the field are often confounded by one of the most counter-intuitive illusions in the field -- the McGurk effect \cite{mcgurk1976}.

Experimental recognition systems for audio are almost always built using phonemes.  There appears to be good agreement as to which phonemes appear in the major languages and what their expected frequency might be.  Once these phonetic units have been recognised then the sequence (together with their probabilities and next-most probable sequence and so on) is fed into a language model which generates hypotheses for words and sentences.  In modern speech recognition language models are powerful and important and have been the subject of decades of work.  There is clearly a huge advantage in a lip-reading system re-using the language model so many lip-reading systems recognise using the visual units, visemes, and then feed the sequence into an acoustic language model modified to cope with visemes.  If visemes exist in the form postulated by linguists e.g.~\cite{jeffers1971speechreading, bozkurt2007comparison}, then there are many choices of visemes.  However there has been surprisingly few examinations of which visemes give the best performance or how fragile that performance is compared to phonetic recognition.

\section{Introduction}
A phoneme is generally regarded as the smallest sound which can be uttered \cite{international1999handbook}. A viseme, which is often said to be the visual equivalent of a phoneme, is not so precisely defined \cite{chen1998audio,fisher1968confusions, Hazen1027972} so we use the working definition: `a viseme is a set of phonemes that have identical appearance on the lips'. Therefore any phoneme falls into one viseme class but a viseme may represent many phonemes: a many to one mapping. 

\begin{figure}[!ht]
\centering
	\includegraphics[width=\textwidth]{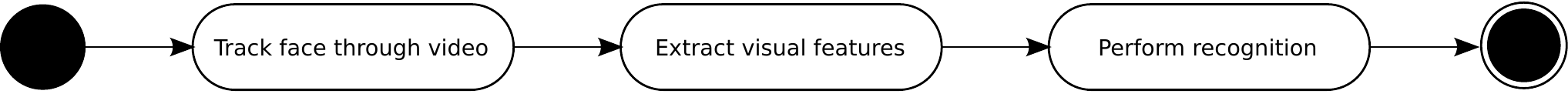} 
	\caption{Steps in a typical lip-reading system}
\label{fig:lip_read}
\end{figure}

A typical lip-reading system is a sequence of tasks as in Figure~\ref{fig:lip_read} and our work is focused within the recognition step. 

Similar to a simplified version of audio recognition whereby we seek to identify a string of unique phonemes~\cite{}, each recognizer is based upon training data of the correctly labelled phoneme. In visual-only recognition we use the same concept of building recognizers based upon visual-only training samples correctly labelled according to a viseme mapping. There is still debate over what the \emph{correct} phoneme-to-viseme mapping is and many have been suggested, e.g \cite{binnie1976visual, fisher1968confusions,kricos1982differences,lip_reading18} but our interest is in the contribution of each viseme to the recognition performance. We look for any particular visemes (or combinations of phonemes) that contribute more to the recognition accuracy. 

We aim to measure the reduction of each unique visemic recogniser in contribution value to the whole task of accurate recognition in continuous speech.
To demonstrate the influence of reduced recogniser classes in visual speech recognition we compare the outputs with those of audio recognition of the same data. For a fair comparison we use the same groupings of phonemes into faux `audio-viseme' recognisers on the audio data. Audio recognition has a higher quantity of classifiers (phonemes) than proposed viseme classes, therefore we hypothesise visual classes have bigger variance in use/purpose towards the whole recognition task. We anticipate, fewer visemes will be used in visual speech recognition than `audio-visemes'  in audio recognition. 

\section{Dataset and feature extraction}
For the first two steps in Figure~\ref{fig:lip_read} we use full face Active Appearance Models (AAMs)~\cite{AAMs} to track the faces through the videos, and lip-only AAMs (one for shape and another for appearance) and using the methods of \cite{Matthews_Baker_2004} we produce two sets of talker-dependent features; shape-only visual features and appearance-only visual features.

\begin{figure}[t]
\centering
\setlength{\tabcolsep}{1pt}
	\begin{tabular}{c c} 
	\includegraphics[width=0.3\textwidth]{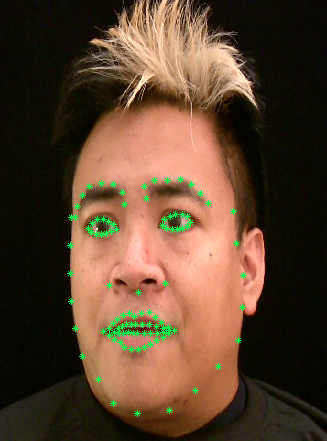} &
	\includegraphics[width=0.3\textwidth]{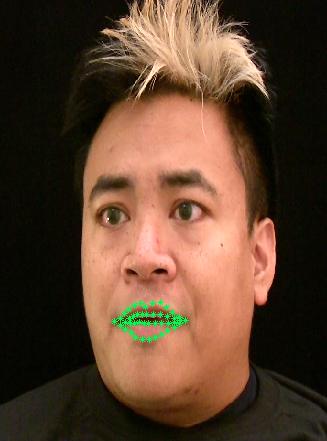} \\
	(a) &	(b) 
	\end{tabular}
\caption{Showing full face shape landmarks for talker T1 (a) and a lip shape landmarks for talker T1 (b).}
\label{fig:mesh}
\end{figure}

\begin{table} [th]
\label{tab:mean_ranks} 
\caption{Frame images from each video.}
\vspace{2mm}
\centerline{
\begin{tabular}{|l|r|r|r|} 
\hline
Video & Num. of AAM train images & Video Length (frames) & Duration\\
\hline  \hline
	Talker1 - 1 & 10 & 21,658 & 00:06:01 \\
	Talker1 - 2 & 10 & 21,713 & 00:06:02 \\
	Talker2 - 1 & 11 & 31,868 & 00:08:52 \\
	Talker2 - 2 & 11 & 33,338 & 00:09:17 \\
\hline
\end{tabular}}
\label{tab:datadata}
\end{table}
Shape features (\ref{eq:shapecombined}) are based solely upon the lip shape and positioning during the duration of the talker speaking e.g. the landmarks in Figure~\ref{fig:mesh}.  The landmark positions can be compactly represented using a linear model of the form:

\begin{equation}
s = s_0 + \sum_{i=1}^ms_ip_i
\label{eq:shapecombined}
\end{equation}
where $s_0$ is the mean shape and $s_i$ are the modes.  The appearance features are computed over pixels, the original images having been warped to the mean shape.  So  $A_0(x)$ is the mean appearance and appearance is described as a sum over modal appearances:
\begin{equation} 
A(x) = A_0(x) + \sum_{i=1}^l{\lambda}_iA_i(x) \qquad \forall x \in S_0
\label{eq:appcombined}
\end{equation} 

The Rosetta Raven data is four videos of recitations of Edgar Allen Poe's poem `The Raven'. There are two talkers, one male, one female. Neither are trained actors and they do not recite the poem with the intended trochaic octameter~\cite{quinn1980critical}. The videos were recorded at $1440\times 1080$ resolution (non-interlaced) at 60 frames per second. Table~\ref{tab:datadata} summarises the video data.

A set of images are extracted from each video (one image per frame) via ffmpeg using image2 encoding at full high-definition resolution ($1440\times1080$).  To construct an initial AAM we select the first frame and nine or ten others randomly. These \emph{training frames} are hand-labelled with a shape model of a face and lips to build a preliminary model for each talker. These models are then fitted, via inverse compositional fitting~\cite{Matthews_Baker_2004} to the remaining frames (Table~\ref{tab:datadata}). Thus we get tracked and fitted full-face talker-dependent AAMs (Figure~\ref{fig:mesh} left) on full resolution lossless PNG frame images (Figure~\ref{fig:lip_read} step 1). 

Next we create a sub-model of only the lips for each talker by decomposing the two full face models (Figure~\ref{fig:mesh} right).
%\begin{figure*}[!htbp]	
%\centering
%\setlength{\tabcolsep}{1pt}
%	\begin{tabular}{c c c} 
%	\includegraphics[width=0.3\textwidth]{figs/image11524.png} & 
%	\includegraphics[width=0.3\textwidth]{figs/image11524-6.png} &
%	\includegraphics[width=0.3\textwidth]{figs/image11524-7.png} \\
%	(a) &	(b) & (c)
%	\end{tabular}
%	\caption{(a) Original resolution ($1440\times1080$) image for T1. (b) T1 downsampled to $60\times45$. (c) T1 restored to  $1440\times1080$.}
%	\label{tab:htkacckey}
%\end{figure*}
From the fitted landmarks, the shape and appearance parameters for each frame are extracted. For talker1 (T1), we retain 6 shape and 14 appearance parameters and for talker2 (T2), 7 shape and 14 appearance parameters. We restrict the feature parameters to retain 95\% of variation from the mean AAM model produced using the whole tracked video data~\cite{AAMs}. (Figure~\ref{fig:lip_read} step 2.)

We did not implement $\Delta\Delta's$ into our extracted features to address co-articulation because we used a phonetic-alignment in the production of our ground-truth benchmark and forced-alignment within the training process of our HMM recognizers.

\begin{table}[h]
\label{visememapping} 
\caption{Phone to viseme mapping.}
\vspace{2mm}
\centerline{
\begin{tabular} {|ll|ll|}
\hline
vID & Phones & vID & Phones \\
\hline  \hline
		v01 & /p/ /b/ /m/ & v10 & /i/ /ih/ \\
		v02 & /f/ /v/ & v11 & /eh/ /ae/ /ey/ /ay/\\
		v03 & /th/ /dh/ & v12 & /aa/ /ao/ /ah/ \\
		v04 & /t/ /d/ /n/ /k/ /g/ /h/ /j/ & v13 & /uh/ /er/ /ax/ \\
		& /ng/ /y/ & v14 & /u/ /uw/\\
		v05 & /s/ /z/ & v15 & /oy/  \\
		v06 & /l/ & v16 & /iy/ /hh/\\
		v07 & /r/ & v17 & /aw/ /ow/  \\ 
		v08 & /sh/ /zh/ /ch/ /jh/ & v18 & silence \\
		v09 & /w/ & &\\
\hline
\end{tabular}}
\end{table}
To have a benchmark for measuring our recognition outputs we produce a ground-truth viseme transcription using the Carnegie Mellon University (CMU) North American pronunciation dictionary\cite{cmudict}, and a word transcription. We convert a phonetic transcript to a viseme transcript assuming 15 visemes, listed in Table~2 which is a combination of Montgomery \textit{et al's} vowel mapping and Walden's consonant mapping\cite{massaro98:talking,walden1977effects}.

\begin{figure}[t]
\centering
	\includegraphics[width=\mywidth,keepaspectratio]{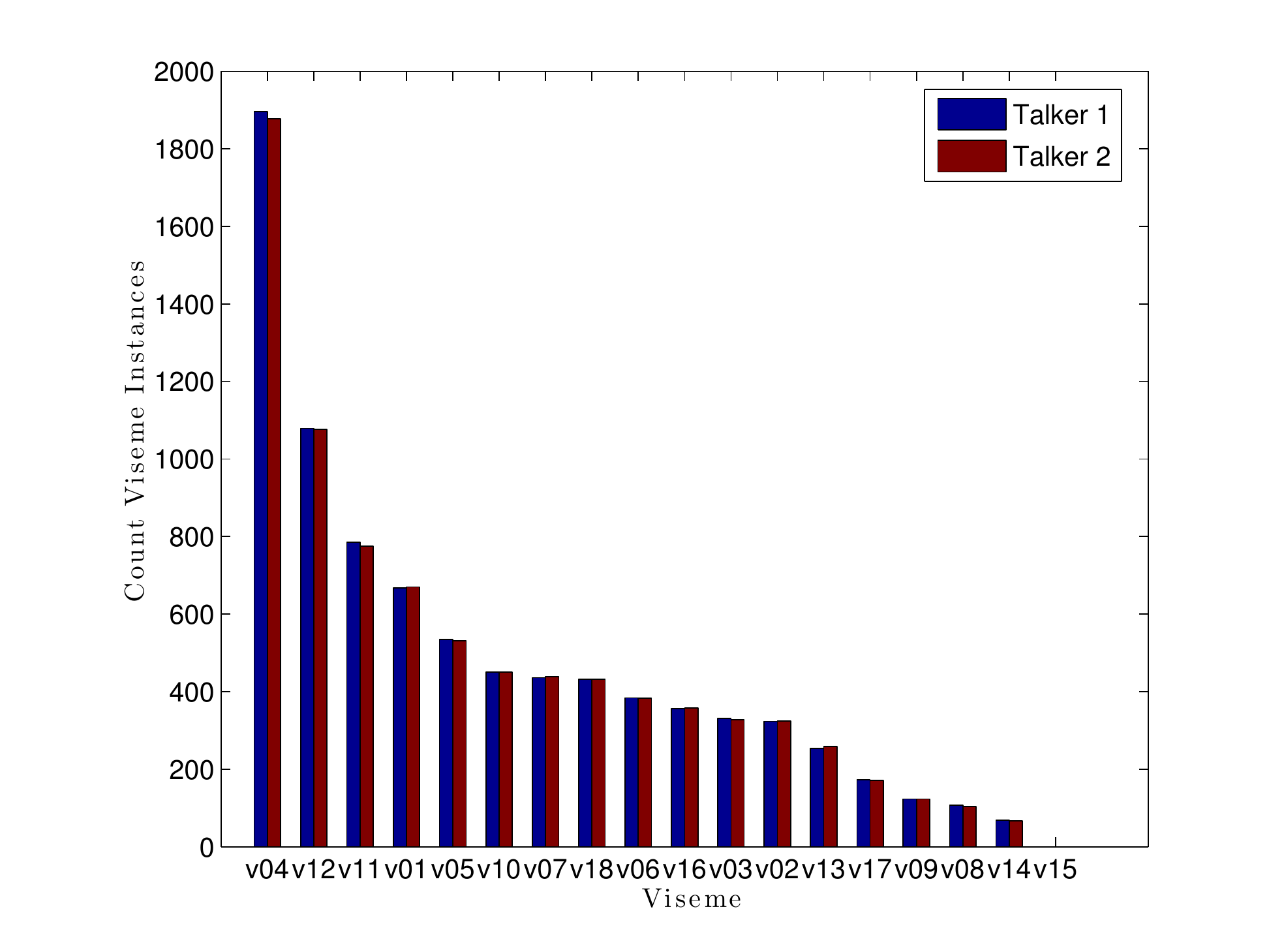} 
	\caption{Viseme counts for both talker transcripts}
\label{fig:viseme_counts}
\end{figure}

\begin{table} [b]
\label{tab:newMap}
\caption{Phone to viseme mapping modified to accomodate restrictions in dataset.}
\vspace{2mm}
\centerline{
\begin{tabular} {|ll|ll|}
\hline
vID & Phones & vID & Phones \\
\hline  \hline
		v01 & /p/ /b/ /m/ & v11 & /eh/ /ae/ /ey/ /ay/ \\
		v02 & /f/ /v/ & v12 & /aa/ /ao/ /ah/ \\
		v03 & /th /dh/ & v13 & /uh/ /er/ /ax/ \\
		v04 & /t/ /d/ /n/ /k/ /g/ /h/ /j/ & v16 & /iy/ /hh/ \\
		& /ng/ /y/ & v17 & /aw/ /ow/  \\
		v05 & /s/ /z/ & v18 & silence  \\
		v06 & /l/ & garb & /u/ /uw/ /oy/ /w/ /sh/ \\
		v07 & /r/ & & /zh/ /ch/ /jh/ \\
		v10 & /i/ /ih/ & &  \\
\hline
\end{tabular}}
\end{table}
The limited availability of large datasets is documented~\cite{cappelletta2012phoneme} so we work within the restrictions of short datasets. Here we note these may not provide adequate training examples of all visemes. Where this happens, we group the untrainable visemes into a single garbage viseme. In this case we select a 150 sample threshold so visemes /v08/, /v09/, /v14/ and /v15/ are grouped. Figure~\ref{fig:viseme_counts} shows the occurrence of visemes listed in Table~\ref{visememapping} in our data and Table~4 shows our revised viseme mapping. 

For each talker, a test fold is randomly selected as 42 of the 108 lines in the poem with replacement. The remaining lines are used as training folds. Repeating this five times gives five-fold cross-validation. Note that visemes cannot be equally represented in all folds. 

For recognition we use Hidden Markov Models (HMMs) implemented in the Hidden Markov Toolkit (HTK)~\cite{htk34}. An HMM is initialised using the `flat start' method using a prototype of five states and five mixture components and the information in the training samples. We choose five states and five mixture components based upon \cite{982900}. We define an HMM for each viseme plus silence and short-pause labels (Table~\ref{visememapping}) and re-estimate the parameters four times with no pruning. 

Next, we use the HTK tool \texttt{HHEd} to tie together the short-pause and silence models between states two and three before re-estimating the HMMs a further two times. Then \texttt{HVite} is used to force-align the data using the word transcript\footnote{We use the \texttt{-m} flag with \texttt{HVite} with the manual creation of a viseme version of the CMU dictionary for word to viseme mapping so that the force-alignment produced uses the break points of the words.}. The HMMs are now re-estimated twice more, however now we use the force-aligned viseme transcript rather than the original viseme transcript used in the previous HMM re-estimations. 

To complete recognition using our HMMs we require a word network as we have a continuous speech dataset. We use \texttt{HLStats} and \texttt{HBuild} to make a Bi-gram Word-level Network (BWN). Finally \texttt{HVite} is used with the network support for the recognition task and \texttt{HResults} gives us both correctness and accuracy viseme recognition values and a viseme confusion matrix for all folds. We have provided the reader with technical details to enable repeatability of our experiments. Please contact the author for original videos. 

\section{Results}
We have extracted figures from the \texttt{HResults} confusion matrices for analysis. For each viseme we have calculated the inverse probability of its recognition $\mbox{Pr}\{v|\hat{v}\}$.  

\begin{figure}[!ht]
\centering
	\includegraphics[width=\mywidth,keepaspectratio]{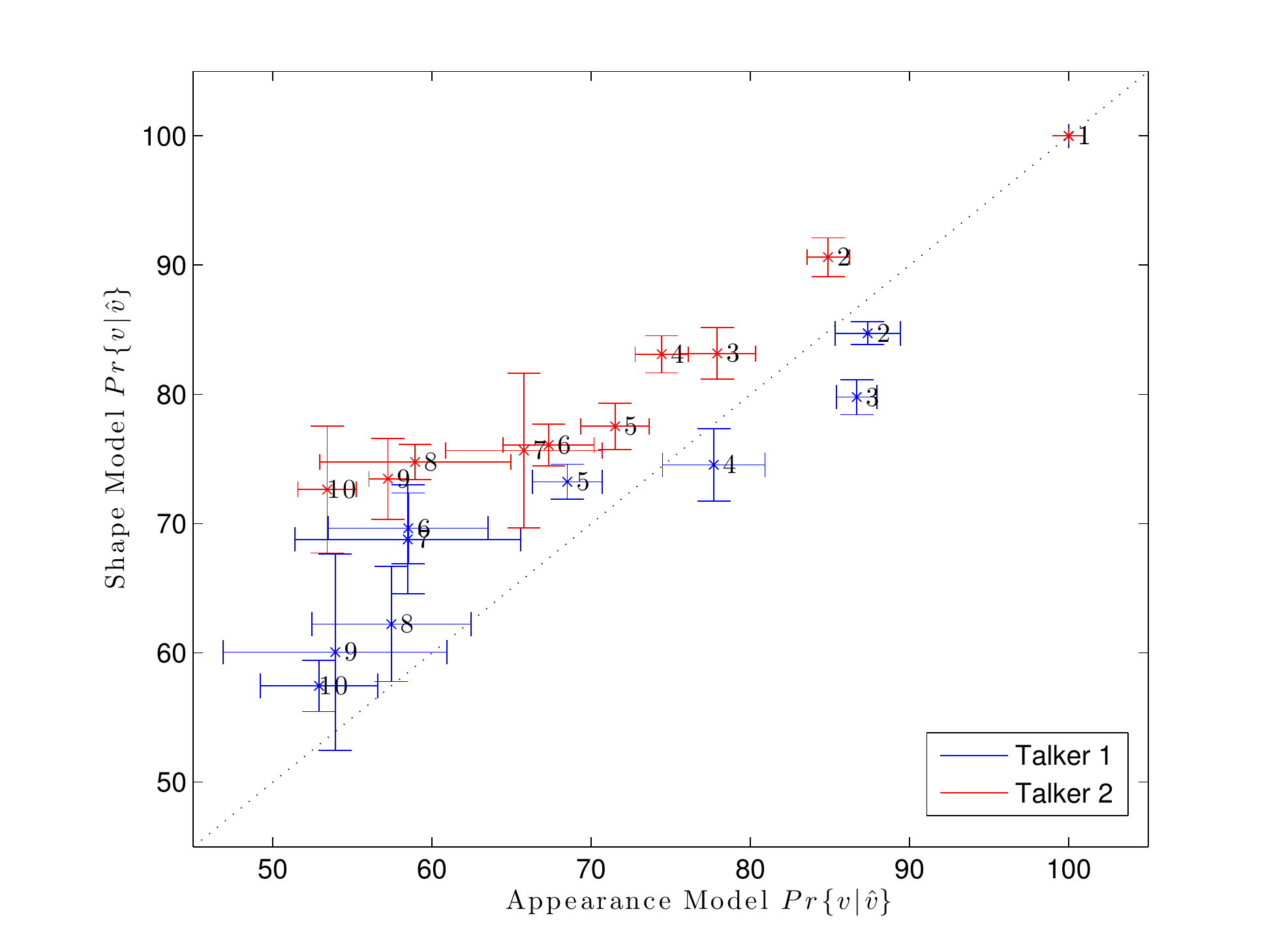} 
	\caption{Relationship between Shape and Appearance model features for both talkers.}
	\label{fig:sVa}
\end{figure}

Figure~\ref{fig:sVa} shows the probability of correct recognition using shape-only features (mean and $\pm 1$ standard error) plotted against the probability of correct recognition using appearance-only features for each viseme. As usual some talkers are better recognised with shape and some with appearance \cite{bowden2013recent}\footnote{The conventional wisdom is that appearance features give the best results but only in studio-type conditions with good tracking.}. Note that the top right-hand point is the visual silence phoneme.  In general, visual silence can be quite variable compared to audio silence because talkers breathe and show emotion. However here, because the source text is a poem, there are well-defined visual silence periods at the start of each line. 

\begin{figure}[!ht]
	\centering
	\includegraphics[width=\mywidth,keepaspectratio]{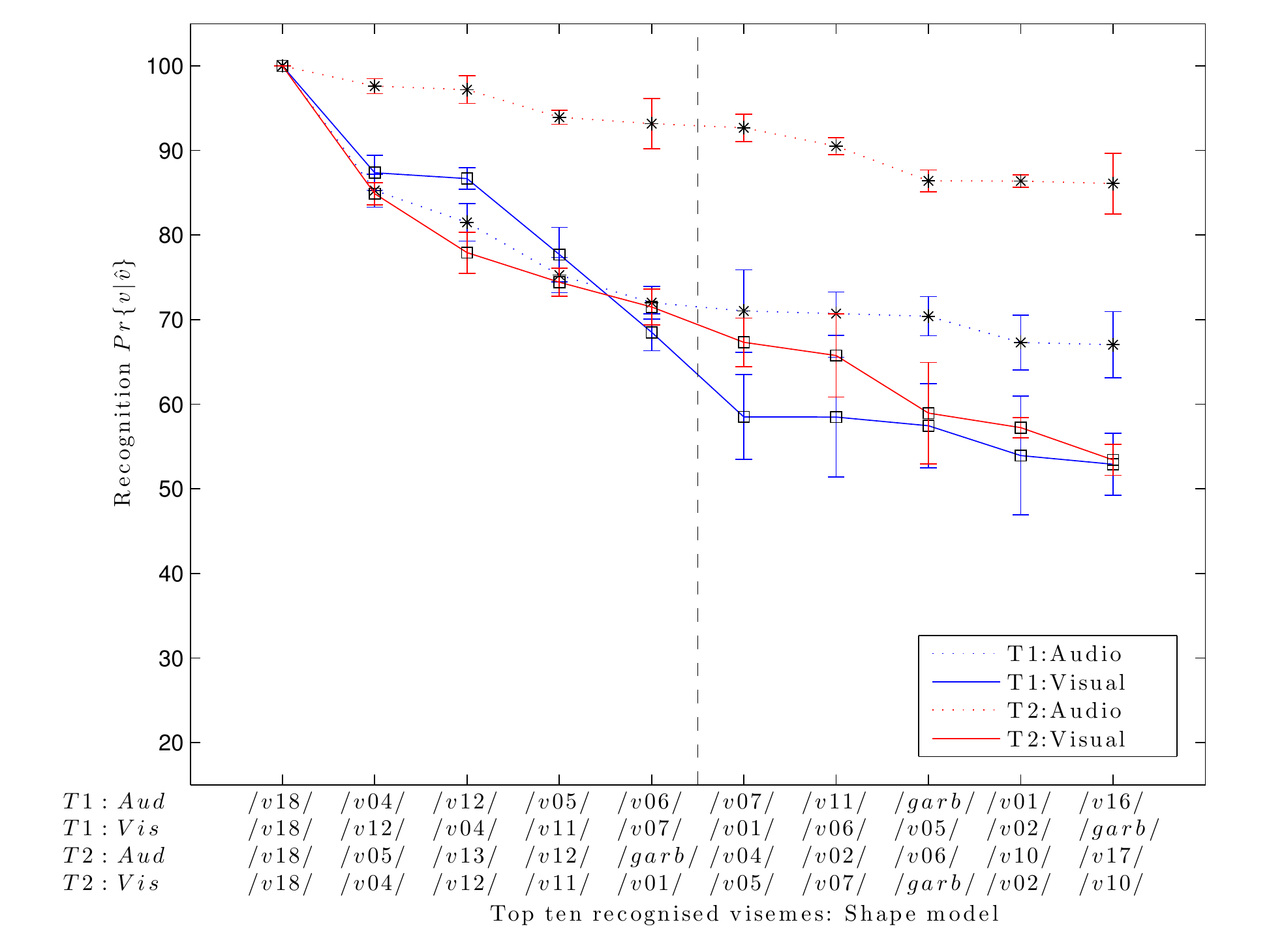} 
	\caption{Top ten viseme recognition probability in descending order with a shape model.}
	\label{fig:shape}
\end{figure}	
\begin{figure}[!ht]
	\centering	
	\includegraphics[width=\mywidth,keepaspectratio]{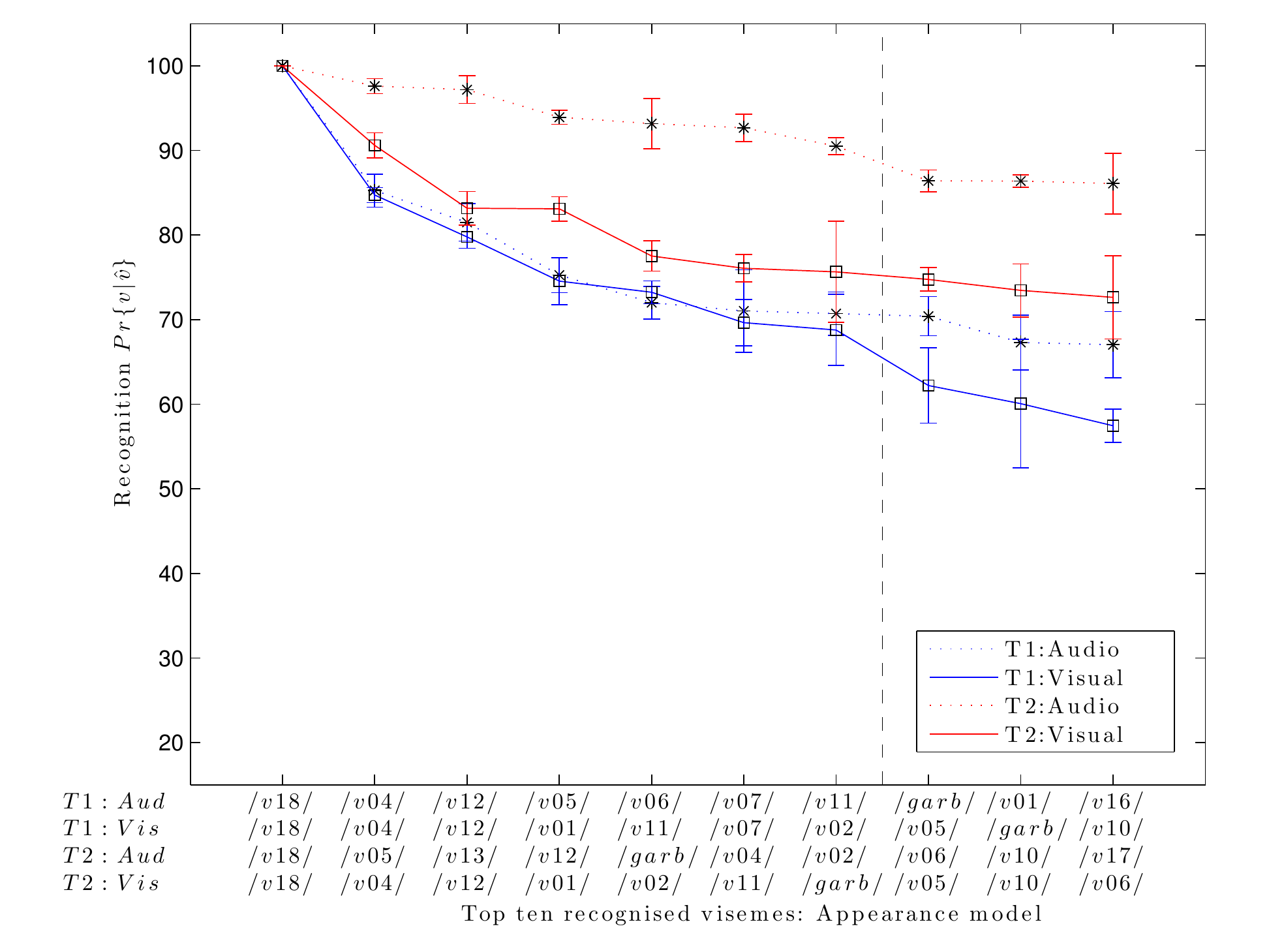} 
	\caption{Top ten viseme recognition probability in descending order with an appearance model.}	
	\label{fig:appearance}
\end{figure}

\begin{table*} [th]
\label{tab:mean_ranks} 
\caption{Ranked mean viseme recognition for Shape, Appearance, Talker 1 and Talker 2.}
\vspace{2mm}
\centerline{
\begin{tabular}{|l|p{14.4cm}|}	
\hline
Feature & Viseme order \\
\hline  \hline
Shape 		& 	/v18/ \{/v04/, /v12\} /v11/ /v01/ /v07/ /v05/ \{/v02/ /v06/ /garb/\} /v10/ \{/v03/ /v13/\} /v16/ /v17/ \\
Appearance 	& 	/v18/ /v04/ /v12/ /v11/ /v01/ /v07/ \{/v02/, /v05/\} /v06/ /garb/ /v10/ /v03/ \{/v13, /v16/\} /v17/  \\
Talker 1 		& 	/v18/ \{/v04/, /v12/\} /v11/ /v01/ /v07/ \{/v02/, /v05/\} /v06/ \{\/v10/, /garb/\} /v03/ /v13/ /v16/ /v17/ \\
Talker 2 		& 	/v18/ \{/v04/, /v12/\} /v11/ /v01/ /v07/ \{/v02/, /v05/\} \{/v06/,/garb/\} /v10/ /v03/ /v13/ /v16/ /v17/ \\
Overall 		& 	/v18/ /v04/ /v12/ /v01/ /v11/ /v07/ \{/v02/, /v05/\} /v19/ /v06/ /v10/ /v13/ /v03/ /v16/ /v17/ \\ 
\hline
\end{tabular}}
\end{table*}

Figures~\ref{fig:shape} and~\ref{fig:appearance} show, for the T1 and T2 shape and appearance models, the probability of correctly recognising the top ten visemes, $\mbox{Pr}\{v|\hat{v}\}$. They also show, the audio performance measured on visemes. The $x$-axis varies by performance; the best performing viseme is on the left hand side which for visual shape and appearance features is silence for all features. 

It has been observed in human lip reading there are few visual cues that are reliable and humans use these combined with rich contextual information to interpret or `fill in the gaps' of what a talker is saying~\cite{erber1975auditory, stork1996speechreading}.  Therefore our hypothesis is that robust audio recognition is based upon a large spread of recognised phones and the resilience in recognition is due to the number of phones contributing to the accuracy. Visually, as with human lip-readers, it is anticipated that fewer visemes would perform the equivalent recognition and, as such, the graph would demonstrate a steeper performance decline over the top performing visemes. 

In Figure~\ref{fig:shape} we do see a greater decline from left to right over the top ten visemes for visual features than for audio for both talkers. We also note that the error bars after the $5^{th}$ position viseme increase, which is consistent with our hypothesis that audio recognition is spread over more visemes to be correct. The top visemes (after silence /v18/) are /v04/, /v12/, /v11/ and /v01/. These are vowels (/v12/, /v11/) and front-of-mouth consonant visemes (/v04/, /v01/). 

Figure~\ref{fig:appearance} demonstrates a shallower decline from left to right than the shape graph in Figure~\ref{fig:shape} but still there is a greater decline for visual features than for audio. The error bars here increase after the $7^{th}$ position viseme\footnote{Note that the order of the audio viseme ordering is identical in both Figures~\ref{fig:shape} and~\ref{fig:appearance} as this is the same experiment.}. The shape of the graph in Figure~\ref{fig:appearance} is similar between audio and video which implies that appearance-based recognition is similar to noisy acoustic recognition for both talkers and hence is less fragile. The top visemes in Figure~\ref{fig:appearance} (not including silence /v18/) are:  /v04/ /v12/ /v11/ /v01/ /v7/ i.e. identical for shape-only in the first six positions.  

Where the error bars increase, we consider this may be due to the small data available, which makes recognition more unreliable due to less well trained HMM classifiers. We have reduced the impact of this with the /garb/ viseme but note with Figure~\ref{fig:viseme_counts} there are similarities between our top performing visemes and those with the most training samples.

\begin{table} [b]
\label{t1vt2} 
\caption{Spearman rank correlation, $r$ and $p$-value for visemes ranked by performance for Talker 1 and Talker 2}
\vspace{2mm}
\centerline{
\begin{tabular}{| l | l | r | r |}
\hline
Talker 1 & Talker 2 & r & p \\
\hline  \hline
Audio & Audio & \underline{0.43} & $1.63\times10^{-2}$ \\
Shape & Shape & \underline{0.92} & 0.00 \\
Appearance & Appearance & \underline{0.93} & 0.00 \\
\hline
\end{tabular}}
\end{table}
Table~\ref{tab:mean_ranks} is the visemes ordered by correctness showing, for example that viseme 18 /v18/ is the best performing viseme overall.  It is natural to ask if the differences in ranking are significant.  To compare the viseme ordering we compute the Spearman rank correlation coefficient, $r$.  The results are shown in Tables~5 and 6.  Also shown is the $p$-value for the null that $r=0$ (randomly ordered).  Those that are significant at the 5\% threshold are underlined.  Talker 2 has poor audio performance which tends to degrade the audio correlation.  Lip-reading does not depend on audio though so these results confirm the strong relation between shape-only and viseme-only classification. Also note for T1 (Figure~\ref{fig:appearance}) the audio ranking is similar to the video ranking although as we have previously noticed there is a more rapid drop-off for video.  
\begin{table} [th]
\label{t1} 
\caption{Spearman rank correlation, $r$ and $p$-values for visemes ordered by feature for Talker 1 (left) and Talker 2 (right) }
\vspace{2mm}
\centerline{
\begin{tabular}{cc}
\begin{tabular}{| l | l | r | r |}
\hline
Talker 1 & Talker 1 & $r$ & $p$ \\
\hline  \hline
Shape & Appearance & \underline{0.90} & 0.00 \\
Audio & Shape & \underline{0.85} & $0.00$ \\
Audio & Appearance & \underline{0.74} & $0.00$ \\
\hline
\end{tabular}
&
\begin{tabular}{| l | l | r | r |}
\hline
Talker 2 & Talker 2 & $r$ & $p$ \\
\hline  \hline
Shape & Appearance & \underline{0.92} & 0.00 \\
Audio & Shape & 0.42 & 0.12 \\
Audio & Appearance & 0.48 & 0.07 \\
\hline
\end{tabular}\\
\end{tabular}}
\end{table}

In Table~\ref{tab:featureStats} we have provided the overall mean and Standard Error Accuracy score for the whole viseme set recognition performance over all five folds. Talker 2 outperforms Talker 1 with all features but for visual features also has a larger degree of error. Appearance features outperform shape for both talkers and audio outperforms appearance for both talkers. As we have seen in Figures~\ref{fig:shape} and \ref{fig:appearance} this recognition is based upon a larger spread of visemes than the shape models with the audio having the largest spread of visemes and hence being the most robust recognition mode.
\begin{table}[h]
\centering
\caption{Mean accuracy scores of each feature type by talker}
	\begin{tabular}{|l|r|c|}
	\hline
	Feature type & Mean & Standard error \\
	\hline \hline
	T1 Audio & 45.6380 & 2.0086 \\
	T2 Audio & 75.8780 & 1.7839 \\
	T1 Shape & 21.7360 & 0.7501 \\
	T1 Appearance & 38.9860 & 0.4637 \\
	T2 Shape & 32.1360 & 1.0339 \\
	T2 Appearance & 52.9540 & 1.9996 \\
	\hline
	\end{tabular}
	\label{tab:featureStats}
\end{table}

\section{Conclusions}
Our principal observations are:
\begin{itemize}
\item Assuming there is enough data to properly train classifiers, then the performance ordering of the visemes is relatively stable across modes of recognition (audio, shape and appearance). 
\item That said, the visual classifiers are far more dependent on the good performance of a few visemes than the audio. 
\item Of the video classifiers, shape is the most dependent on a few visemes. 
\end{itemize}

These are important results because they illuminate the often made observation that lip-reading is fragile. In other words if one cannot build classifiers for a few critical visemes then lip-reading is impossible. In a human lip-reading context, humans are often trained to recognise a small number of critical gestures which are then processed via a very sophisticated language and context model to create a transcript. 

In audio is it surprisingly rare to see this effect measured even though a good acoustic unit will have accuracies that are at least 10\% higher than an average unit (the mean audio viseme performance on T2 is 76\% for the whole viseme set). 

It is important to acknowledge that most work in this field focuses on improving mean accuracies over the set of all visemes which can conceal the real source of overall performance. A system that achieves a mean viseme accuracy of, say 53\% maybe one that contains one supremely accurate viseme classifier or it maybe a system that has a set of classifiers of much more modest performance. 

This paper therefore raises two different tactics for improving lip-reading systems. Either one makes the best viseme classifiers better or, one focuses upon improving the worst. At this stage we do not know which tactic is likely to be more successful but we hope this methodology allows future work to focus attention where it is likely to do the most good. 

\bibliography{icpr2014}   %>>>> bibliography data in report.bib
\bibliographystyle{spiebib}   %>>>> makes bibtex use spiebib.bst

\end{document}